\documentclass[sigconf,square,sort,comma,numbers,compress]{acmart}

\AtBeginDocument{%
  \providecommand\BibTeX{{%
    \normalfont B\kern-0.5em{\scshape i\kern-0.25em b}\kern-0.8em\TeX}}}

\setcopyright{acmcopyright}
\copyrightyear{2018}
\acmYear{2018}
\acmDOI{10.1145/1122445.1122456}

\usepackage{tikz}

\usepackage[boxed,ruled]{algorithm2e}

\usepackage{mathtools}

\usepackage{multicol}
\usepackage{multirow} 
\usepackage{wrapfig}
\usepackage{subfig}
\usepackage[boxed,ruled]{algorithm2e}

\usepackage{amsmath}
\usepackage{subfig}
\usepackage{diagbox}

\usepackage{natbib}
\usepackage{color}

\DeclareMathOperator*{\argmin}{argmin} 

\begin{document}

\title{Zero-Space Cost Fault Tolerance for Transformer-based Language Models on ReRAM}




\author{Bingbing Li}
\email{bingbing.li@uconn.edu}
\affiliation{%
  \institution{University of Connecticut}
  \country{USA}
}
\author{Geng Yuan}
\email{geng.yuan@uga.edu}
\affiliation{%
  \institution{University of Georgia}
  \country{USA}
}
\author{Zigeng Wang}
\email{zigeng.wang@uconn.edu}
\affiliation{%
  \institution{University of Connecticut}
  \country{USA}
}
\author{Shaoyi Huang}
\email{shaoyi.huang@uconn.edu}
\affiliation{%
  \institution{University of Connecticut}
  \country{USA}
}
\author{Hongwu Peng}
\email{hongwu.peng@uconn.edu}
\affiliation{%
  \institution{University of Connecticut}
  \country{USA}
}
\author{Payman Behnam}
\email{payman.behnam@gatech.edu}
\affiliation{%
  \institution{Georgia Institute of Technology}
  \country{USA}
}
\author{Wujie Wen}
\email{wwen2@ncsu.edu}
\affiliation{%
  \institution{North Carolina State Univeristy}
  \country{USA}
}
\author{Hang Liu}
\email{hang.liu@rutgers.edu}
\affiliation{%
  \institution{Rutgers}
  \country{USA}
}
\author{Caiwen Ding}
\email{caiwen.ding@uconn.edu}
\affiliation{%
  \institution{University of Connecticut}
  \country{USA}
}

\begin{abstract}


Resistive Random Access Memory (ReRAM) has emerged as a promising platform for deep neural networks (DNNs) due to its support for parallel in-situ matrix-vector multiplication. However, hardware failures, such as stuck-at-fault defects, can result in significant prediction errors during model inference. While additional crossbars can be used to address these failures, they come with storage overhead and are not efficient in terms of space, energy, and cost. In this paper, we propose a fault protection mechanism that incurs zero space cost. Our approach includes: 1) differentiable structure pruning of rows and columns to reduce model redundancy, 2) weight duplication and voting for robust output, and 3) embedding duplicated most significant bits (MSBs) into the model weight. We evaluate our method on nine tasks of the GLUE benchmark with the BERT model, and experimental results prove its effectiveness.

\end{abstract}




\vspace{-6mm}
\keywords{ReRAM, fault tolerance, compression, Transformer, BERT}


\maketitle


\vspace{-3mm}
\section{Introduction}

Resistive Random Access Memory (ReRAM), a non-volatile memory based on programmable resistance, provides both data storage and in-situ dot product calculations to accelerate DNN inference. Several ReRAM-based neural accelerators, such as PRIME~\cite{chi2016prime} and ISAAC~\cite{shafiee2016isaac}, have been proposed for energy-efficient applications.


\begin{figure}[t]
\centering
\includegraphics[width=0.99\columnwidth]{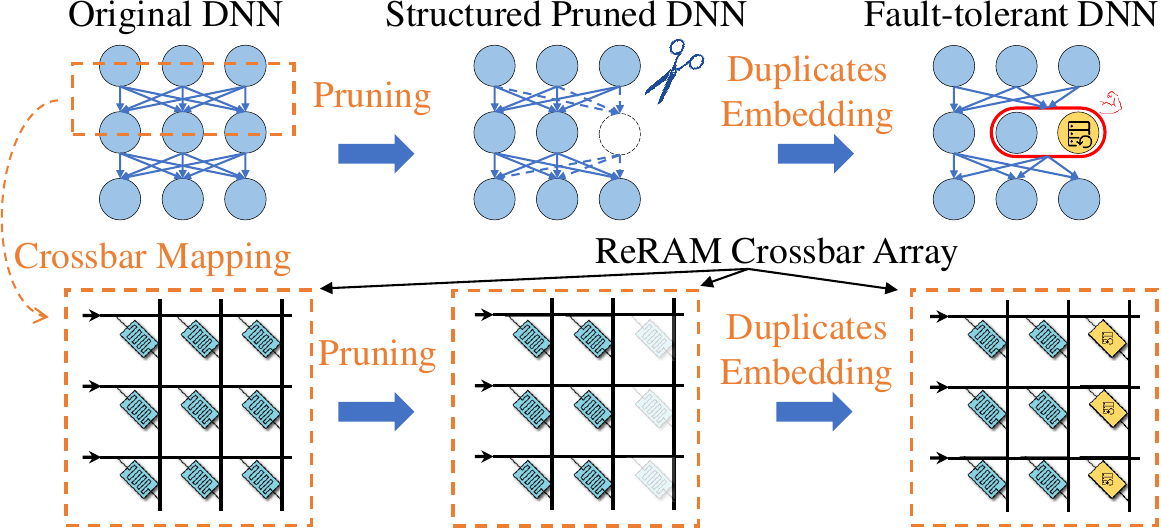}
\vspace{-3.5mm}
\caption{Zero-space Cost Fault Tolerance on ReRAM.}
\label{fig:structure_prune}
\vspace{-6mm}
\end{figure}

However, due to the immaturity of fabrication technology, ReRAM-based systems can suffer from possible failures. Permanent faults, known as hard faults, cause the resistance states of ReRAM cells to be fixed at high resistance (stuck-at-0 fault) or low resistance (stuck-at-1 fault), due to fabrication defects~\cite{chen2014rram}.
In the case of deep neural network inference, the impact of hardware failure is amplified as errors in the front layers accumulate and propagate layer by layer during forward propagation. 
The situation becomes worse for the application of large scale natural language processing (NLP) models, such as BERT~\cite{devlin2018bert}, GPT~\cite{radford2019language}, etc.
Before applying these models on ReRAM platforms, the weight parameters are quantized~\cite{park2017weighted} and binarized into bit matrices, consisting of the most significant bits (MSBs) and least significant bits (LSBs).
Among these bit matrices, the MSBs have the most significant impact on model performance. Stuck-at-1 failures, which occur more than 5 times as frequently as stuck-at-0 failures, can increase parameter values by 10 or even 100 times, leading to significant degradation in model performance.
In this paper, we focus on MSB failure tolerance.

 Different methods are proposed to mitigate the effects of hardware failures~\cite{chen2014rram,liu2015vortex,yan2022computing,yan2022swim,yan2023improving,wang2022fault}. Liu~\cite{liu2015vortex} proposed an on-device training method by generating a dedicated network that can tolerate faults in the memoristor chip distribution. Chen~\cite{chen2014rram} proposed a ``write-verify" scheme that involves conducting online tests and repairs regularly to address faults. In order to achieve a balance between robustness and accuracy on ReRAM, Wang~\cite{wang2022fault} proposed a fault-tolerant training scheme to enhance resilience to stuck-at-fault defects. However, these approaches introduce storage overhead and require additional ReRAM modules and auxiliary ADC/DAC circuits. This raises the question: Can we improve the robustness of the model on ReRAM without incurring storage overhead? Fortunately, the answer is yes. By leveraging model pruning and compression techniques, we can obtain a sparse model structure and save a significant number of ReRAM modules and ADC/DAC circuits.


DNN networks are commonly acknowledged to have a significant level of redundancy. To address this issue and ensure model accuracy, pruning methods such as deep compression~\cite{han2016deep_compression} and heuristic pruning~\cite{dai2019nest} as well as quantization methods like fixed-point~\cite{lin2016fixed} and binarization~\cite{hubara2016binarized} have been proposed and utilized. Structured pruning methods, such as efficient pruning~\cite{li2020efficient} and FTrans~\cite{li2020ftrans}, have also been proposed to improve the efficiency of hardware and have shown significant acceleration gains on actual hardware. However, these methods rely on manual parameter selection and fine-tuning based on human experience. Additionally, they require customized debugging for different DNN models, such as adopting different sparsity expectations for different layers~\cite{zhang2018systematic}. In order to reduce the need for manual intervention, Xiao~\cite{Xiao2019AutoPruneAN} proposed automatic pruning, which automatically finds the optimal sparse model structure by incorporating trainable auxiliary parameters.


Our motivation for this research is twofold: 1) given that some parameters are both crucial and prone to errors, why not create backups for these parameters to improve the overall model's robustness? 2) since DNNs contain a significant amount of redundancy, can't these redundancies be leveraged to compensate for the additional hardware introduced by the backups?







In this paper, we propose a fault tolerance strategy that does not introduce additional parameter space. We present an automatic structured (specifically, row and column) pruning method to reduce the parameter space and use these pruned parameter space to store important backup parameters. Furthermore, we introduce a weight voting strategy to enhance the robustness of the entire network.
%
%
Our contributions in this paper can be summarized as follows:
\leftmargini=4mm
\begin{itemize}

\item We propose a novel structured differentiable pruning method, the row and column differentiable pruning, to achieve row and column sparsity in the model automatically. We use gate parameters to determine which rows or columns to retain and update both the gate and weight parameters to achieve a better balance between model accuracy and sparsity. 

\item Based on the observation that most of the parameter values are smaller than the half of the maximal value of the corresponding layer, we introduce a bit-flipping voting strategy to leverage the duplicated parameters to recover the original output and improve the final model's performance.

\item We propose an MSB embedding method to eliminate the storage overhead introduced by the duplicated parameters. Specifically, we adopt an in-place redundant duplicates storage strategy by embedding these duplicated MSBs into the model weights during weight-crossbar mapping on ReRAM.




\end{itemize}

\section{Background and limitation}
\subsection{ReRAM-based systems: Advantage and Limitations}
Deep Neural Networks (DNNs) have become dominant in the field of artificial intelligence due to their high accuracy and scalability.
As DNN models have grown in size, they present great challenges for conventional hardware due to the increasing computational cost and memory storage requirements.
However, the weights and computing units are separated in the conventional Von Neumann architectures, which results in significant data movements~\cite{9499856}.

Resistive Random Access Memory (ReRAM), compared to conventional accelerators, has shown superior performance in terms of inference speed boosting, extremely low energy comsumption, and in-situ computation. This is attributed to the fact that ReRAM crossbars can naturally reduce data movements and computation costs, thus accelerating DNNs.

Despite the high hardware efficiency enjoyed by ReRAM-based Computation Systems (\textbf{RCS}), ReRAM circuits are still prone to hardware faults due to immature fabrication technology. These faults can be categorized into hard faults and soft faults. Soft faults include Read-One-Disturb (R1D) and Read-Zero-Disturb (R0D), while the most common types of hard faults are Stuck-At-Faults (SAF), which include Stuck-At-Zero (SA0) and Stuck-At-One (SA1)~\cite{6725492}. In the case of soft faults, an ReRAM cell can still function since its resistance can change. However, for hard faults, the previous solution is not applicable anymore due to the unchangeable resistance. Consequently, weights cannot be programmed into the faulty cell. This leads to a significant drop in accuracy since the matrix stored in a crossbar is incomplete. The inference accuracy is sensitive to the number of defective
memristors~\cite{10.1145/3287624.3287707,9283554}.

\subsection{Pruning for efficient ReRAM utilization}

In the pruning strategy, magnitude-based pruning methods such as ADMM~\cite{zhang2018systematic} and Reweighted L1~\cite{zhang2021unified} can reduce redundant model parameters by assigning the expected sparsity for each layer. However, this approach often suffers from sub-optimal results, as the pruning performance relies solely on the researchers' experience. To address this limitation, various efficient one-shot automatic pruning algorithms have been proposed, in which the pruned model structure and weights are jointly learned through back-propagation. Another solution for sparse model structure searching is reinforcement learning. However, this method requires extensive training time due to the large search space and relies heavily on the researcher's experience to achieve stable training performance.

Differentiable pruning, on the other hand, assigns auxiliary parameters to determine the weight retention and updates these parameters using back-propagation. This approach can achieve a high compression ratio while maintaining model accuracy. Unlike earlier methods, differentiable pruning does not rely on layer-wise sparsity design and can determine the layer sparsity automatically. Straight-through estimators (STEs)~\cite{srinivas2017training, Xiao2019AutoPruneAN} are introduced in pruning to learn discrete sparse network structures. These estimators assign overridden coarse gradients to binarization functions. Differentiable neural network architecture search with Gumbel Softmax~\cite{liu2018darts} and the use of Gumbel Softmax in attention head pruning~\cite{voita2019analyzing} for transformer models have also been proposed. However, neither of these algorithms has implemented differentiable pruning schemes for structured pruning to enhance hardware computation efficiency.

\section{Fault-tolerated ReRAM}
To improve the hardware efficiency and robustness of SAF, we propose a fault-tolerated ReRAM implementation method that involves: 1) leveraging differentiable structured pruning to remove redundant weights while preserving model accuracy, 2) introducing weight voting strategy by leveraging the duplicated bits for fault tolerance of the Most Significant Bits (\textbf{MSBs}) to maintain model accuracy, 3) 
introducing an in-place duplicates storage strategy by embedding these duplicated bits into the model weights during weight-crossbar mapping.

\begin{figure}[t]
\centering
\includegraphics[width=0.65\columnwidth]{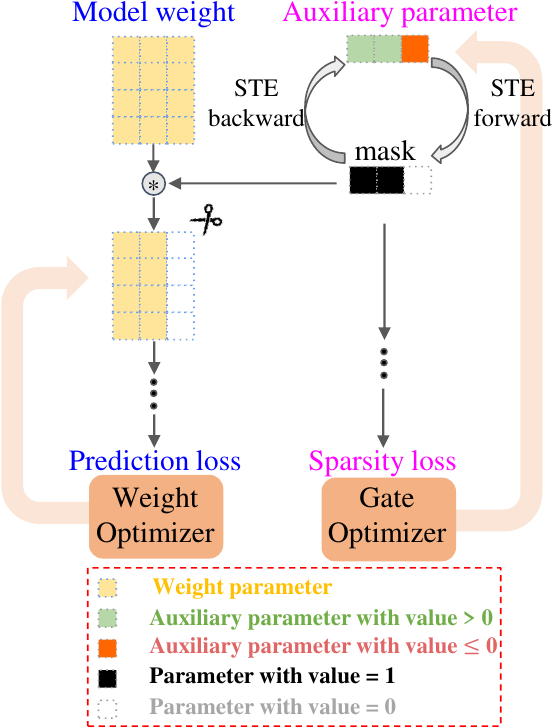}
\vspace{-3.5mm}
\caption{Differentiable structured (e.g. column) pruning framework.}
\label{fig:structure_prune}
\vspace{-6mm}
\end{figure}

\subsection{Stage 1: Differentiable structured pruning}
Structured pruning is performed to reduce model redundancy by dropping entire weights in different rows or columns of weight matrix, thus achieving a row- or column-sparse model architecture.

For an $N$-layer DNN model, the weight parameters of the $k$-th layer are denoted as $W_k$. The pruning problem can be formulated based on $L0$ regularization:

\begin{equation}\label{eq:weight_pruning_org} 
    \argmin_{W} f(X,\{W_k\}_{k=1}^{N}) + \mu \sum_{k=1}^{N}||\{W_k\}_{k=1}^{N}||_{l_0},
\end{equation}
\noindent where $f(*)$ denotes the accuracy loss function representing the model prediction loss, $X$ represents the model input, and $\mu$ represents the penalty factor of model sparsity.

To overcome the optimization problem of $L0$ regularization and prune the weights in a back-propagation manner, we introduce a indicator function, $H(*)$, to represent the pruning status of the model weights. For structured pruning, we formulate the row pruning function, $H_{row}$, and column pruning function, $H_{col}$, as follows:

\begin{eqnarray}
\small
\label{eq:row_column_pruning}
    H_{row}(W_k;M_k)=\mathop{Concat}\limits_{i}(M_k\otimes W\left[i,:\right],dim=0)\\
    H_{col}(W_k;M_k)=\mathop{Concat}\limits_{j}(M_k^T\otimes W\left[:,j\right],dim=1)\\
    M_k\left[i\right]=
    \begin{cases}
      \mathbf{0}, & \text{if} \ W\left[i,:\right] \text{or } W\left[:,i\right] \text{ is pruned}; \\   
      \mathbf{1}, & \text{otherwise}.
    \end{cases}
\end{eqnarray}

\noindent where $Concat(*)$ represents the concatenating function to concatenate different rows or columns according to $dim$ selection, $M_i$ is a column vector and represents the binary mask for the structure (row or column), $W\left[i,:\right]$ represents the $i$-th row of the weight matrix and $W\left[:,j\right]$ represents the $j$-th column, $\otimes$ represents the Kronecker product~\cite{roger1994topics}.

Then the structured pruning problem can be reformulated as the following optimization problem:

\begin{equation}\label{eq:structured_pruning} 
    \argmin_{W,M} f(X,\{H(W_k;M_k)\}_{k=1}^{N}) + \mu \sum_{k=1}^{N}||\{M_k\}_{k=1}^{N}||_{l_1}
\end{equation}

To make it differentiable in Eq.\ref{eq:structured_pruning}, we employ learnable discrete functions called straight
through estimators (STEs)~\cite{hubara2016binarized,Xiao2019AutoPruneAN} $g$ to describe the mask M
and thus $M$ can be formulated as:

{\small    
\begin{equation} \label{eq:aux_parameter} 
M_k\left[i\right]=g(W'_{k}\left[i\right]) =
    \begin{cases}
      0, & \text{if} \ W'_{k}\left[i\right] \leq 0 \\   
      1, & \text{if} \ W'_{k}\left[i\right] > 0.
    \end{cases}
\end{equation}}

\noindent where $W'_{k}$ is the auxiliary parameter to \textit{control} the \textit{open} and \textit{close} of the binary masking gate $M_{k}$, and $M_{k}$ is represented as a step function $g$ with a continuous auxiliary parameter $W_{k}'$. 

Finally, we formulate the structured pruning problem as:



\begin{align}
\label{eq:structured_pruning_final}
  \argmin_{W,W'} \mathcal{L}_t 
  & = \argmin_{W,W'} \left[ f\{X, H(W_k;g(W_{k}'))\}_{k=1}^{N} \right. \\
  & \left. + \mu \sum_{k=1}^{N} || g(W_{k}')_{k=1}^{N} ||_{l_1} \right. \Big]
\end{align}

\noindent where $\mathcal{L}_t$ is the total loss during structured pruning.

The problem described in Eq.~\ref{eq:structured_pruning_final} is a mixed integer programming problem. We design two optimzers: weight optimizer to update retained model weight using normal back-propagation and gate optimizer to update auxiliary parameter as shown in Fig~\ref{fig:structure_prune}.
The binary gates are non-differentiable and we introduce coarse gradients~\cite{hubara2016binarized}
to calculate the gradient of the binarization function.
%
We use Softplus STE in~\cite{Xiao2019AutoPruneAN} and the auxiliary parameter $W'$ can be updated as:

{\small  
\begin{equation}\label{eq:gate_param_update} 
    \begin{aligned}
    {W'} & \xleftarrow{} {W'} - l_{rg} * \frac{\partial \mathcal{L}_{t}}{\partial W'} \\
    \end{aligned}
\end{equation}}\noindent
\begin{equation}
\small
\label{eq:gate_param_update2}
    \frac{\mathcal{L}_{t}}{\partial W'} =
    \frac{\partial \mathcal{L}_{t}}{\partial {M}} *
    \frac{\partial {M}}{\partial W'} =
    \frac{\partial \mathcal{L}_{t}}{\partial {M}} *
    Softplus({W'})
\end{equation}
where $l_{rg}$ is the learning rate of the gate optimizer to update $W'$.

\begin{figure}[b]
\centering
\includegraphics[width=0.9\columnwidth]{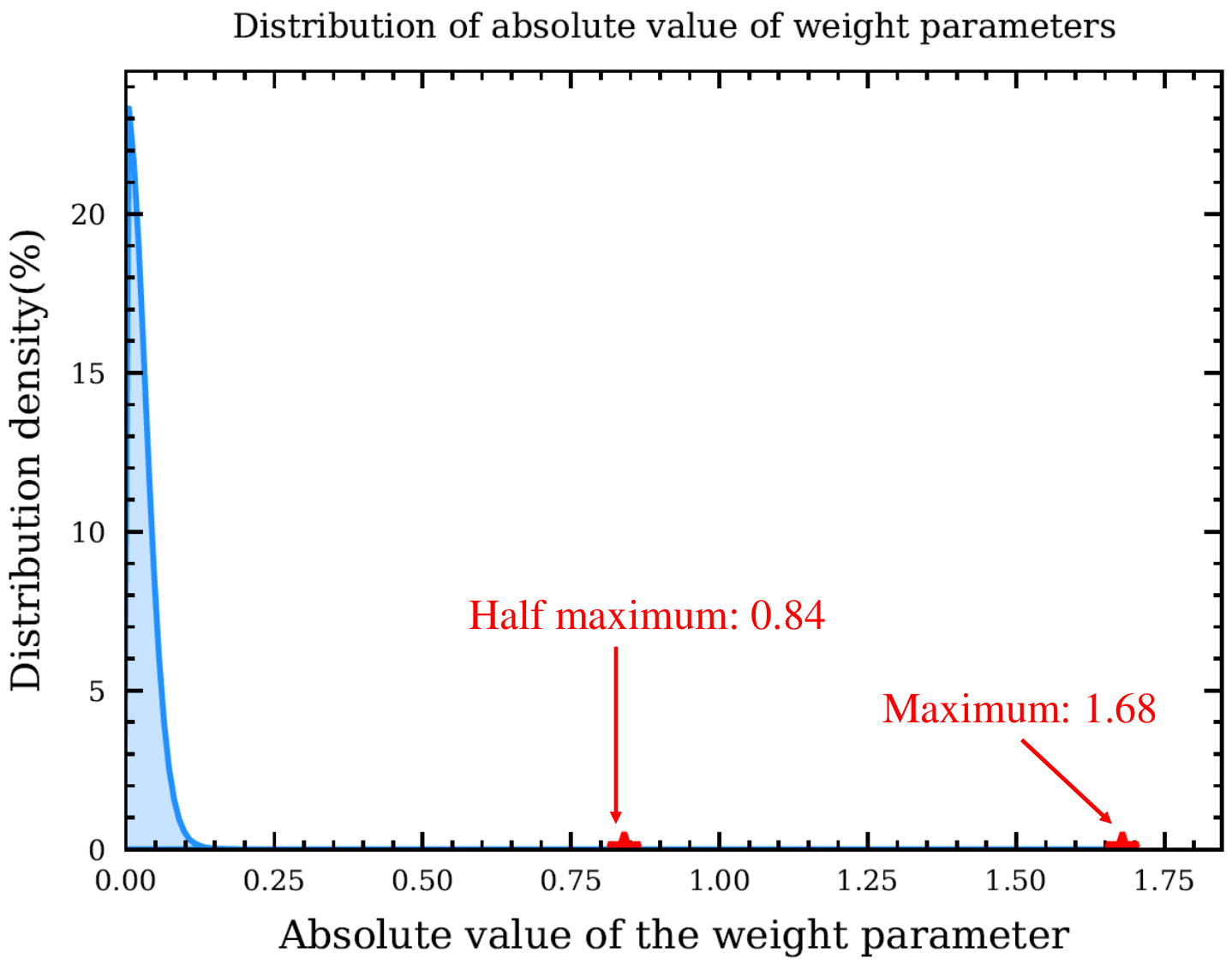}
\vspace{-3.5mm}
\caption{Weight distribution of the 1st layer weight matrix of BERT model: most values of the parameters are much smaller than the half of the weight maximum.}
\label{fig:weight_distri}
\vspace{-6mm}
\end{figure}

\subsection{Stage 2: Fault tolerance}
The occurrence of hard faults can lead to significant degradation in model performance. In the case of single-bit implementation of ReRAM-based systems, the normal fault ratio can exceed 0.05\%. Furthermore, the occurrence of a Stuck-At-Faults (SAF) on the most significant bits (MSBs) can result in a weight value change of up to 10$\times$ or even 100$\times$, which is disastrous during model inference. Figure~\ref{fig:weight_distri} shows the weight distribution of the first encoder layer in BERT, where the maximum absolute value is 1.68, while more than 99\% absolute weight values are smaller than half of the maximum. This illustrates the high risk of significant value change in model weights with MSB SA1 fault since more than 99\% of MSBs values are zero.

\begin{figure}[t]
\centering
\includegraphics[width=1.0\columnwidth]{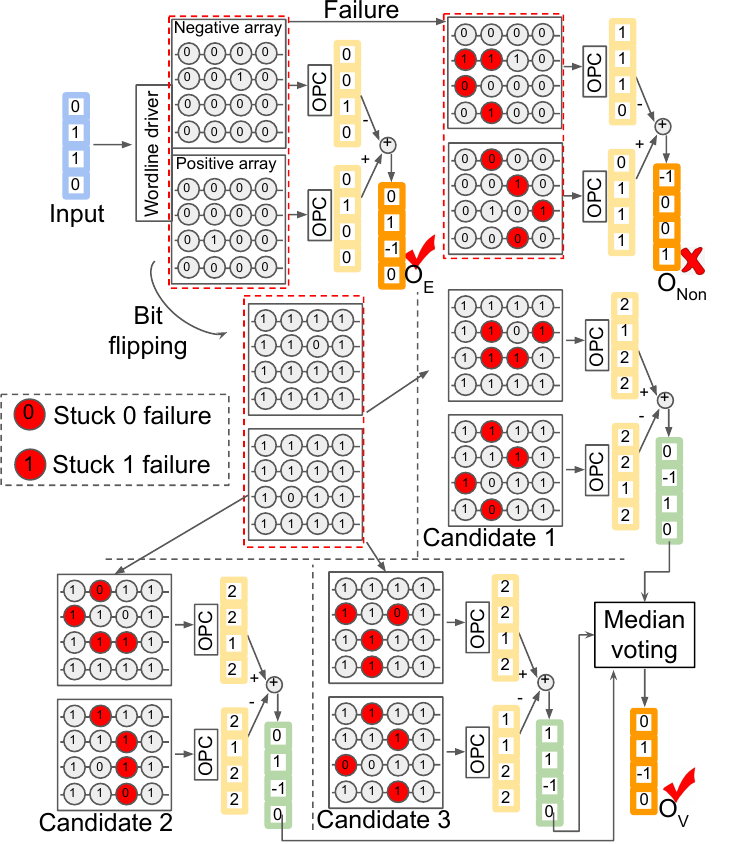}
\vspace{-3.5mm}
\caption{Duplication and output voting on ReRAM crossbars ($O_E$ is the expected output, $O_{Non}$ is the output without voting, $O_V$ is the output after voting, OPC module is the Output Peripheral Component).}
\label{fig:duplicate_vote}
\vspace{-6mm}
\end{figure}

\begin{figure}[b]
\centering
\vspace{-5mm}
\includegraphics[width=0.75\columnwidth]{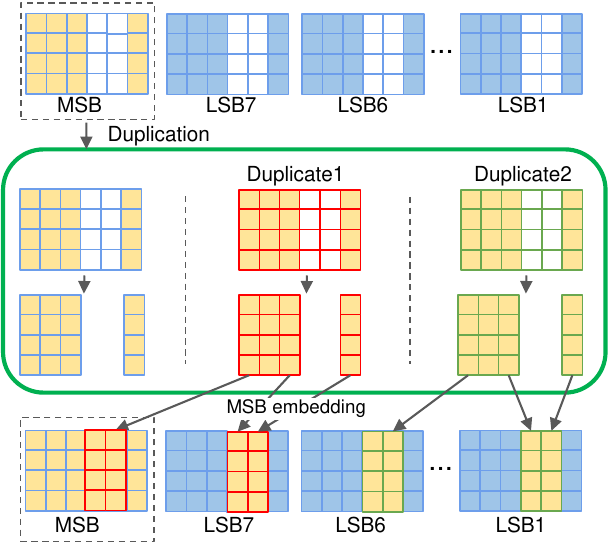}
\vspace{-3.5mm}
\caption{MSB embedding to eliminate storage overhead.}
\label{fig:MSB_embed}
\vspace{-6mm}
\end{figure}

To improve fault tolerance, we first duplicate MSBs while mapping model weights to ReRAM crossbars.
During inference on ReRAM, we use these duplicates as candidates to vote for the final circuit output. Additionally, to leverage the weight parameter distribution finding, we design a bit-flipping voting strategy by mapping flipped bits into the ReRAM platform and changing the sign of the obtained intermediate results. This is due to two reasons: 1) since more than 99\% model parameter values are smaller than the half of the maximum, mora than 99\% MSBs are zero, which is severely sensitive to SAF; 2) the frequency of SA1 failure is much larger than SA0~\cite{beckmann2016nanoscale}.


\subsubsection{Quantization and binarization}
In order to facilitate hardware implementations, we employ $n$-bit equal-distance quantization, resulting in a total of $V = 2^{n-1}$ quantization levels. Specifically, for each layer weight $\mathbf{W}_k$, we obtain the quantized weight matrix, $\mathbf{W_q}$, by quantizing the weights into a set of quantization values \{-$\frac{V}{2}q_i$,...,-$q_i$,$q_i$,...,$\frac{V}{2}q_i$\}, where $q_i=max(abs(\mathbf{W}_k))/(2^{n-1})$.

The quantized weights are then binarized as follows:

{\small
\begin{equation}
\begin{tiny}
\mathbf{W_q}=\mathbf{Sign}*(\mathbf{MSB}*2^{n-1}+\mathbf{LSB}_{n-1}*2^{n-2}+...+\mathbf{LSB}_{1}*1)
\end{tiny}
\end{equation}
}

\noindent where $\mathbf{Sign}$ is the sign matrix of $\mathbf{W}$, $\mathbf{MSB}$ represents the most significant bit (\textbf{MSB}) matrix of $\mathbf{W_q}$, and $\mathbf{LSB}$ represents the less significant bit (\textbf{LSB}) matrix of $\mathbf{W_q}$. During ReRAM crossbar mapping, the $\mathbf{Sign}$ matrix is not directly mapped to the crossbars. Instead, we use a negative ReRAM array and a positive ReRAM array to represent the $\mathbf{Sign}$ matrix.

\subsubsection{MSB duplication and result voting}
After quantization and binarization of the model, we obtain binary weights consisting of MSBs and LSBs. Since MSBs has the most significant impact on model performance, we duplicate the MSB matrices and design the output voting strategy to enhance the resilience to SAF defects on ReRAM.

\begin{algorithm}[t]
\scriptsize
\captionsetup{font={small}}
  \caption{Model inference with MSB fault tolerance for each layer}
  \label{alg:MSB_fault_tolerate}
   \KwIn{weight matrix $\mathbf{W}$, layer input $\mathbf{X}$, number of candidates $\mathbf{T}$}
   \KwOut{model layer output $\mathbf{O}$}
  Do n-bit quantization and binarization of $\mathbf{W}$ to obtain $\mathbf{B}=[\mathbf{MSB},\mathbf{LSB_{n-1}},...,\mathbf{LSB_{1}}]$;\\
  Duplicate and bit-flip $\mathbf{MSB}$ $\mathbf{T}$ times to obtain $\mathbf{MSB}_1$, ..., $\mathbf{MSB_T}$ ;\\
  \ForEach{duplicate $\mathbf{MSB}_k$}
    {Add SA0 and SA1 failure noise into $\mathbf{MSB}_i$ to simulate SAF;\\
    Calculate the candidate output of $\mathbf{MSB}_i$ as $\mathbf{O_{MSB_i}}$ by using OPCs and summing;\\
    }
    Do median voting using $[\mathbf{O_{MSB_1}}, ..., \mathbf{O_{MSB_T}}]$ to derive the final MSB array output $\mathbf{O_{MSB}}$;\\
  Calculate the output of 
  $[\mathbf{LSB_{n-1}}, ...,\mathbf{LSB_{1}}]$
  as $[\mathbf{O_{LSB_{n-1}}},...,\mathbf{O_{LSB_{1}}}]$;\\
  Use $[\mathbf{O_{MSB}}, \mathbf{O_{LSB_{n-1}}}, ...,\mathbf{O_{LSB_{1}}}]$ to derive the final output $O$.\\
\end{algorithm}

\begin{algorithm}[b]
\scriptsize
\captionsetup{font={small}}
  \caption{ReRAM crossbar mapping with zero space cost for one layer}
  \label{alg:MSB_embedding}
   \KwIn{sparse $\mathbf{MSB}_1$, sparse $\mathbf{LSBs}$, additional $\mathbf{MSBs}$ ($\mathbf{MSB}_2$, $\mathbf{MSB}_3$, ..., $\mathbf{MSB}_K$), column pruning index $\mathbf{Index}$}
   \KwOut{Mapped crossbar arrays with embedded MSBs}
  Set additional bit matrix $\mathbf{A}$ as an empty list;\\
  \ForEach{each $\mathbf{MSB}_k (k=2,...,K)$}{
    \ForEach{each column of $\mathbf{MSB}_k$}{
        \If{column index is not in $\mathbf{Index}$}
            {append column into $\mathbf{A}$}
    }
  }
  Distribute $\mathbf{A}$ into zero columns of $\mathbf{MSB}_1$ and $\mathbf{LSBs}$.
\end{algorithm}

\begin{table*}[h]\tiny
	\centering
	\caption{{Differentiable structured pruning among the nine GLUE benchmark tasks.}}\label{table:pruning_results}
    \vspace{-3.5mm}
	\resizebox{1\textwidth}{!}{
		\begin{tabular}{l| l l l l l l l l l}
			\hline
			Models & MNLI & QQP & QNLI & SST-2 & CoLA & STS-B & MRPC & RTE & WNLI\\
\hline
\textbf{BERT$_{\mathrm{BASE}}$ \cite{devlin2018bert}} & 84.6 & 91.2 & 90.5 & 93.5 & 52.1 & 85.8 & 88.9 & 66.4 & - \\
\textbf{BERT$_{\mathrm{BASE}}$ (ours)} & 83.9 & 91.4 & 91.1 & 92.7 & 53.4 & 85.8 & 89.8 & 66.4 & 56.3 \\
\textbf{BERT$_{\mathrm{BASE}}$ prune (ours)} & 83.19 & 90.86 & 90.72 & 92.75 & 54.72 & 85.73 & 89.74 & 66.79 & 56.34\\
\textbf{Sparsity} & 36.2\% & 30.03\% & 31.53 & 40.13\% & 31.60\% & 30.42\% & 31.35\% & 42.22\% & 45.64\%\\
\hline
		\end{tabular}
	}
\end{table*}

Before voting, we use a 0-1 flipped bit matrix for model weight and crossbar mapping. To store the weight sign, negative and positive arrays are used to store the negative and positive values of the model weight. Instead of directly mapping the weight bit value to the crossbars, we invert the bit value between 0 and 1, and map the inverted bit value into the negative and positive crossbar arrays.

During inference, we calculate the candidate outputs by changing the sign of the output of the Output Peripheral Components (OPCs) before summing, and use the median value of the three candidate results as the final layer output. The entire process is illustrated in Figure~\ref{fig:duplicate_vote} and Algorithm~\ref{alg:MSB_fault_tolerate}. In our test, we manually add SA0 and SA1 faults to simulate the ReRAM SAF by giving different hardware failure rate.



\subsection{Stage 3: Embedding MSB candidates for weight-crossbar mapping}
To address the storage overhead introduced by MSBs duplication, we adopt an in-place redundant duplicates storage strategy by embedding these backup MSBs into the model weights during mapping.

\begin{figure}[t]
\centering
\includegraphics[width=0.9\columnwidth]{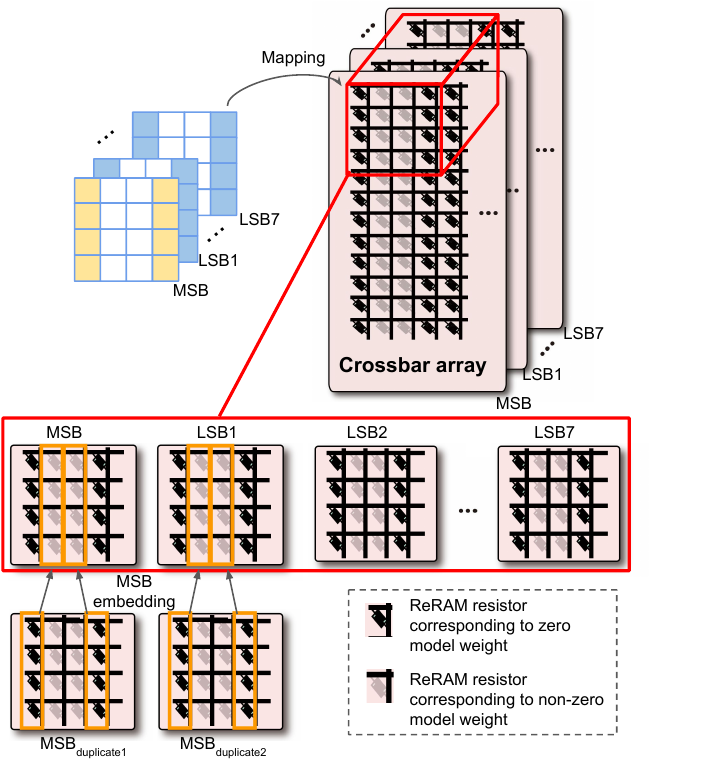}
\vspace{-3.5mm}
\caption{Duplicated MSBs embedding}
\label{fig:crossbar_map}
\vspace{-6mm}
\end{figure}

The process of inserting additional MSB columns into the positions of zero columns after structured pruning is depicted in Figure~\ref{fig:MSB_embed} and Algorithm~\ref{alg:MSB_embedding}. In the case of 8-bit quantization, if the sparsity (percentage of pruned columns) exceeds 30\%, there is no additional storage overhead introduced. Figure~\ref{fig:crossbar_map} showcases the weight-crossbar mapping in real ReRAM modules, where the duplicated MSBs columns are mapped to the positions corresponding to zero model weights. As described in Section 4.2, the derived sparse models have more than 30\% sparsity on all nine datasets, demonstrating the eliminated storage overhead.

\section{Evaluation}

\subsection{Experiment settings}
Our baseline model is the BERT-base-uncased model from HuggingFace \cite{wolf2020transformers}. To evaluate the performance of the pruned model, we use the GLUE benchmark, which consists of 9 tasks covering different categories of NLP tasks with different degrees of difficulty and dataset scales.
We report accuracy scores for SST-2, QNLI, RTE, and WNLI; Matthews Correlation Coefficient (MCC) for CoLA; F1 scores for QQP and MRPC; and Spearman correlations for STS-B.
%
For BERT, we use the official BERT$_{\mathrm{BASE}}$ uncased model as our pre-trained model. This model has 12 layers ($L$ =12; hidden size $H$ = 768; self-attention heads $A$ = 12), with total number of parameters of 110 million. We use the same fine-tuning hyperparameters as described in \cite{devlin2018bert}.


\subsection{Results of differentiable structured pruning}
We compared the performance of our pruned models with the baseline models (unpruned models) and summarized the results in Table~\ref{table:pruning_results}. As mentioned in Section 3.2.2, we consider sparsities larger than 30\% to meet the MSB embedding requirements. Across all nine tasks, we achieved more than 30\% sparsity with either no or less than 1\% drop in accuracy. Surprisingly, our pruned models even showed accuracy improvements on the QNLI, CoLA, and MRPC datasets, which could be attributed to the reduction of redundant weights that led to decreased overfitting. Moreover, our models successfully met the requirement of eliminating the introduced storage overhead for all tasks in the GLUE benchmark.

\subsection{Distribution of weight parameters}
To gain a better understanding of the weight parameter distribution and further contribute to the mapping and voting improvement, we analyzed the weight distribution of word embedding layer and all 12 encoder layers of the BERT model. Table~\ref{table:weight_distri} presents the distribution of the absolute values of the model parameters in different layers. The maximum value represents the highest value within the entire layer and we define the \textit{large parameters} as the parameters with the values larger than the half of the maximum of the corresponding weight matrix.
We counted the number of large parameters and observed that more than 99\% of the weight parameters were smaller than half of the maximal value of the weight matrix. Consequently, more than 99\% of the MSB elements are zeros. This proves the effectiveness of the bit-flipping strategy during weight-crossbar mapping.

\begin{table*}[htb]\tiny
	\centering
	\caption{Distribution of the absolute values of the model weight paramters (WB: Word Embedding layer; E: Encoder layer)}\label{table:weight_distri}
 \vspace{-3.5mm}
	\resizebox{1\textwidth}{!}{
		\begin{tabular}{l| l l l l l l l l l l l l l}
			\hline
			& WB & E1 & E2 & E3 & E4 & E5 & E6 & E7 & E8 & E9 & E10 & E11 & E12\\
\hline
\#param(million) & 23.44 & 2.36  & 2.36  & 2.36  & 2.36  & 2.36  & 2.36  & 2.36  & 2.36  & 2.36  & 2.36  & 2.36  & 2.36  \\
param maximum & 0.95 & 1.68 & 2.15 & 3.33 & 6.64 & 6.82 & 6.02 & 4.81 & 3.56 & 3.13 & 3.08 & 6.37 & 2.15 \\
\#(large param) & 7 & 6 & 18 & 2 & 3 & 1 & 1 & 1 & 2 & 4 & 14 & 2 & 5 \\

\hline
		\end{tabular}
	}
\end{table*}

\begin{figure}[htb]
\centering
\includegraphics[width=0.9\columnwidth]{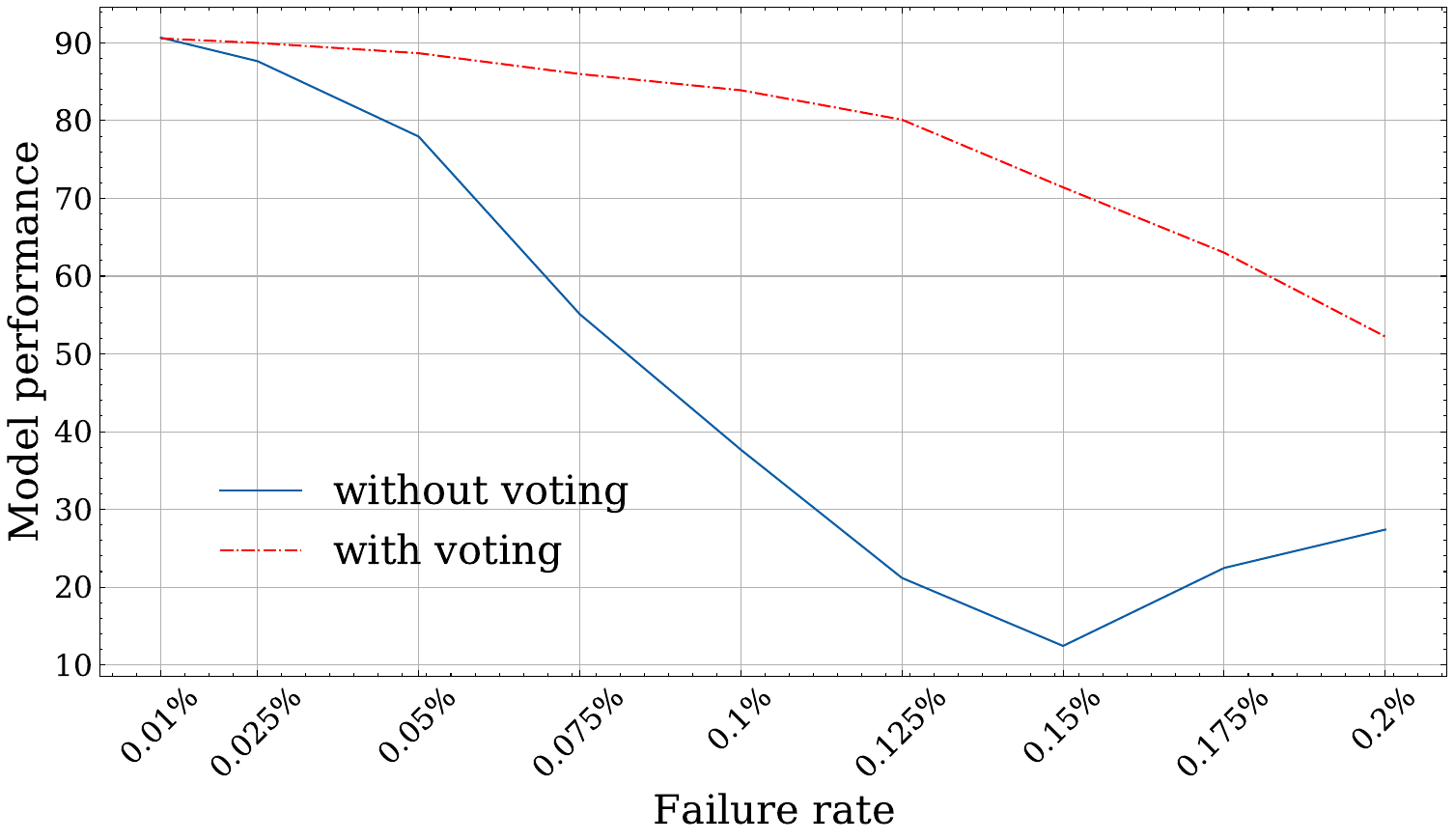}
\vspace{-3.5mm}
\caption{Model performance with different failure rates with and without output voting on MRPC.}
\label{fig:vote_performance}
\vspace{-6mm}
\end{figure}

\begin{table*}[htb]\tiny
	\centering
	\caption{{Model performance with different failure rates with and without failure tolerance with BERT on MRPC dataset.}}\label{table:voting_results}
    \vspace{-3.5mm} 
	\resizebox{1\textwidth}{!}{
		\begin{tabular}{ll| l l l l l l l l l }
			\hline	\multicolumn{2}{c|}{\diagbox{Method}{Failure rate}} & 0.01\% & 0.025\% & 0.05\% & 0.075\% & 0.1\% & 0.125\% & 0.15\% & 0.175\% & 0.2\% \\
            \hline
            \multirow{2}*{Without voting} & Accuracy Mean & 90.68	& 87.65	& 77.96	& 55.11	& 37.68	& 21.18	& 12.44	& 22.47	& 27.40 \\
            & Accuracy Variance & 0.36	& 0.97	& 14.35	& 169.49	& 286.78	& 227.09	& 178.24	& 231.09	& 269.88 \\

\hline
\multirow{2}*{With voting (\textbf{ours})}
& Accuracy Mean & 90.58	& \textbf{90.26}	& \textbf{88.68}	& \textbf{86}	& \textbf{83.91}	& \textbf{80.13}	& \textbf{71.43}	& \textbf{63.02}	& \textbf{52.23} \\

& Accuracy Variance & \textbf{0.20}	& \textbf{0.50} &	\textbf{1.20}	& \textbf{2.33} &	\textbf{4.85}	&	\textbf{10.23}	&	\textbf{50.74}	&	\textbf{93.22}	&	\textbf{155.73} \\

\hline
		\end{tabular}
	}
 \vspace{-3mm}
\end{table*}

\subsection{MSB candidates voting}
As described in Section 3.2.2, we simulate ReRAM failures (SA0 and SA1 failures) by introducing 0 or 1 noise. We set the overall failure ratio to follow the widely-used SA0 and SA1 ratio of 1.75:9.04~\cite{beckmann2016nanoscale,chen2014rram}. Figure~\ref{fig:vote_performance} displays the model accuracy with and without candidate voting when different percentages of ReRAM failures are introduced. The results demonstrate the significant improvement in accuracy preserving with voting. With a 0.05\% failure rate, our voting method achieves an accuracy improvement of nearly 10\% over the baseline model without voting at the same failure rate. Moreover, our method can tolerate 2.5 times the failure rate compared to the baseline, while maintaining a drop in accuracy below 10\%.

\section{Conclution}
\vspace{-1mm}

In this paper, we propose a fault tolerance strategy without introducing additional parameter space on ReRAM. Differentiable structured pruning method is proposed to automatically reduce model redundancy and a fault-tolerant model inference method based on bit-flipping, most significant bits (MSBs) duplication, and result voting to enhance the capability to handle Stuck-at-fault (SAF) defects on ReRAM. Additionally, we embed the MSBs duplicates into the pruned zero parameter space to eliminate the storage overhead. Experimental results show that our pruning method can remove enough parameter space to eliminate the storage space overhead (larger than 30\%) of the introduced duplicates, while the model performance is maintained with either no or less than 1\% accuracy drop on all nine GLUE datasets. For the SAF defects, the proposed method can tolerate 2.5 times the failure rate with the same model accuracy as the baseline, which proves the effectiveness to enhance the resilience to ReRAM SAF defects.

\vspace{-3mm}

 \newpage
 
\bibliography{ref}










\end{document}